\documentclass[10pt,twocolumn,letterpaper]{article}

\usepackage{cvpr}              

\usepackage{graphicx}
\usepackage{amsmath}
\usepackage{amssymb}
\usepackage{booktabs}

\usepackage{multirow}
\usepackage{makecell}
\usepackage{appendix}

\usepackage[pagebackref,breaklinks,colorlinks]{hyperref}

\usepackage[capitalize]{cleveref}
\crefname{section}{Sec.}{Secs.}
\Crefname{section}{Section}{Sections}
\Crefname{table}{Table}{Tables}
\crefname{table}{Tab.}{Tabs.}

\def\cvprPaperID{2544} 
\def\confName{CVPR}
\def\confYear{2023}

\begin{document}

\title{Discriminator-Cooperated Feature Map Distillation for GAN Compression}

\author{Tie Hu$^{1}$, Mingbao Lin$^3$, Lizhou You$^1$, Fei Chao$^{1}$, Rongrong Ji$^{1,2}$\thanks{Corresponding Author}\\
$^1$MAC Lab, School of Informatics, Xiamen University \\
$^2$Institute of Artificial Intelligence, Xiamen University\\
$^3$Tencent Youtu Lab\\
{\tt\small \{hutie, lmbxmu, youlizhou\}@stu.xmu.edu.cn, \{fchao, rrji\}@xmu.edu.cn}
}

\maketitle

\begin{abstract}
Despite excellent performance in image generation, Generative Adversarial Networks (GANs) are notorious for its requirements of enormous storage and intensive computation. As an awesome ``performance maker'', knowledge distillation is demonstrated to be particularly efficacious in exploring low-priced GANs.
In this paper, we investigate the irreplaceability of teacher discriminator and present an inventive discriminator-cooperated distillation, abbreviated as DCD, towards refining better feature maps from the generator. 
In contrast to conventional pixel-to-pixel match methods in feature map distillation, our DCD utilizes teacher discriminator as a transformation to drive intermediate results of the student generator to be perceptually close to corresponding outputs of the teacher generator. 
Furthermore, in order to mitigate mode collapse in GAN compression, we construct a collaborative adversarial training paradigm where the teacher discriminator is from scratch established to co-train with student generator in company with our DCD.
Our DCD shows superior results compared with existing GAN compression methods. For instance, after reducing over 40$\times$ MACs and 80$\times$ parameters of CycleGAN, we well decrease FID metric from 61.53 to 48.24 while the current SoTA method merely has 51.92. This work's source code has been made accessible at \url{https://github.com/poopit/DCD-official}.
\end{abstract}

\section{Introduction}
\label{sec:intro}

%
Image generation transforms random noise or source-domain images to other images in user-required domains. Recent years have witnessed the burgeoning of generative adversarial networks (GANs) that lead to substantial progress in image-to-image translation~\cite{isola2017image, zhu2017unpaired, chen2018cartoongan, choi2018stargan}, style transfer~\cite{gatys2016image, gatys2015texture, xu2021drb}, image synthesis~\cite{karras2019style, karras2020analyzing, brock2018large, radford2015unsupervised, zhang2019self}, \emph{etc}.
Image generation has a wide application in daily entertainment such as TikTok AI image generator, Dream by WOMBO, Google Imagen, and so on.
Running platforms performing these applications are typically featured with poor memory storage and limited computational power.
However, GANs are also ill-famed for the growing spurt of learnable parameters and multiply-accumulate operations (MACs), raising a huge challenge to the storage requirement and computing ability of deployment infrastructure.

\begin{figure}[!t]
  \centering
  \includegraphics[width=0.45\textwidth]{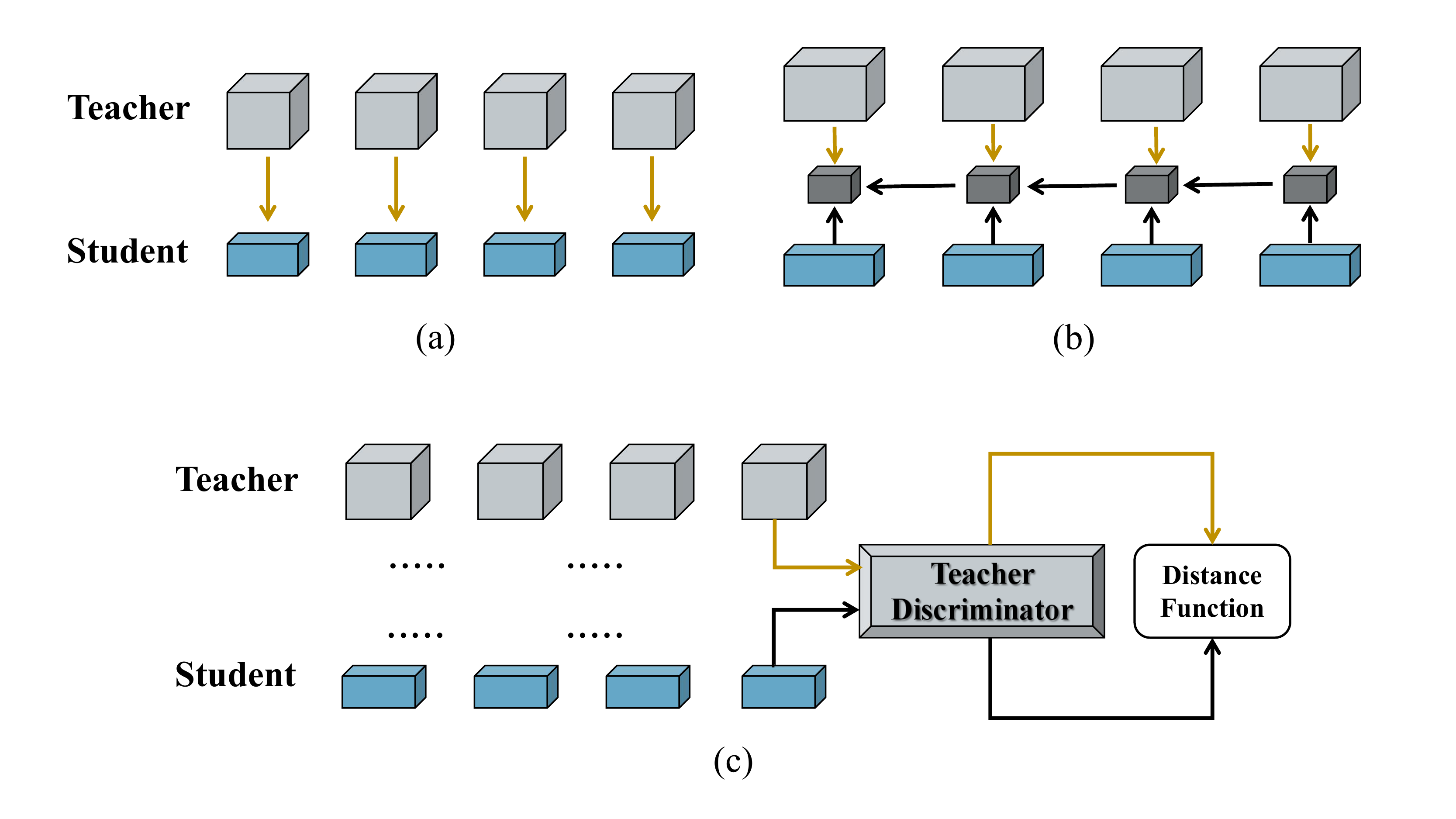}
  \vspace{-1.0em}
  \caption{(a) Layer-by-layer feature map distillation~\cite{romero2014fitnets}. (b) Cross-layer feature map distillation~\cite{chen2021distilling}. (c) Our discriminator-cooperated feature map distillation.}
  \label{method_review}
  \vspace{-1.0em}
\end{figure}


%
%
To address the above dilemma for better usability of GANs in serving human life, methods such as pruning~\cite{li2021revisiting, ren2021online, li2022learning, chen2021gans}, neural network architecture search (NAS)~\cite{fu2020autogan, jin2021teachers, li2020gan} and quantization~\cite{wang2020gan, wang2019qgan}, have been broadly explored to obtain a smaller generator. On the premise of these compression researches, knowledge distillation, in particular to distilling feature maps, has been accepted as a supplementary means to enhance the performance of compressed generators~\cite{li2020gan, hou2021slimmable, aguinaldo2019compressing, wang2018kdgan, chen2020distilling, li2020semantic}. Originated from image classification, as illustrated in Fig\,\ref{method_review}(a), feature map based distillation, which extracts information of intermediate activations and transfers the knowledge from the teacher model to the student one, has been extensively explored and demonstrated to well improve the capability of lightweight models~\cite{romero2014fitnets, chen2021wasserstein, zagoruyko2016paying, li2022knowledge, yang2022masked}. Distinctive from passing on common feature maps from teacher to student, AT~\cite{zagoruyko2016paying} calculates feature attentions as the delivered knowledge; MGD~\cite{yang2022masked} randomly masks feature maps to indirectly guide the student to learn from the teacher;
KRD~\cite{chen2021distilling} uses a cross-layer distillation method to allow the ``new knowledge" of the student to learn from the ``old knowledge" in teacher, as shown in Fig.\,\ref{method_review}(b). Whatever, most methods execute pixel-to-pixel feature maps matching between teacher and student.

Alike to the implementations on image classification, feature map based distillation is also considered in GAN compression. For example, DMAD~\cite{li2021revisiting} considers a well pre-trained discriminator to absorb high-level information from the teacher-generated image, and fuses it with intermediate activations from the teacher generator, results of which are passed to the corresponding position of the student generator. %
OMGD~\cite{ren2021online} utilizes an online multi-granularity strategy to allow a deeper teacher generator and a wider one to simultaneously deliver output image knowledge of different granularities to the student generator.
%
These two methods follow the pipeline of image classification to tune the intermediate outputs of student generator with those of teacher generator in a fashion of per-pixel matching.
Although the sustainable progress on multiple benchmark datasets demonstrates the efficacy of intermediate activation outputs, the feature-based distillation, as we reveal in this paper, is not well compatible with the very nature of generating perceptually similar images and adversarial training paradigm in GANs.


%
Concretely speaking, conversely to image classification that relies on feature vector representations, the essence of image generation is to improve perceptually alike between the real images and generated images. Two important facts cause it is eventually difficult to use per-pixel match to analyze a pair of images: First, two similar images can contain many different pixel values; Second, two dissimilar images can still comprise the same pixel values. Thus, it is not suitable to simply use the per-pixel match.
Regarding adversarial training in GANs, a generator learns to synthesize samples that best resemble the dataset, meanwhile a discriminator differentiate samples in the dataset from the generator generated samples. The adversarial results finally lead the generator to creating images of out-of-the-ordinary visual quality, indicating that the discriminator is also empowered with informative capacity and can be exploited to enrich the distillation of feature maps.
Therefore, it might be inappropriate to directly extend feature map distillation in image classification to image generation. And GAN compression oriented feature map distillation with discriminator included remains to be well explored.

In order to achieve this objective, in this paper, we propose a discriminator-cooperated distillation (DCD) method to involve the teacher discriminator in distilling feature maps for student generator. A simple illustration is given in Fig.\,\ref{method_review}(c), in contrast to the vanilla pixel-wise distance constraint, our DCD measures the distance at the end of teacher discriminator with the intermediate generator outputs as its inputs. Our DCD is perspicacious in multiple benchmark datasets with a simple implementation. Akin to perceptual loss~\cite{johnson2016perceptual} which employs a pre-trained neural network such as a VGG model~\cite{simonyan2014very} to extract features upon which the $\ell_1$ distance is calculated from activations of hidden layers, the teacher discriminator in DCD also acts as a feature extractor. Due to pooling operations in the hidden layers, feature maps from different sources (student generator and teacher generator) as inputs to the discriminator may lead to identical latent representations, therefore encouraging natural and perceptually pleasing results.
In addition, the proposed DCD is used in conjunction with collaborative adversarial training, which is also simple but perspicacious to allow the student generator to fool the discriminator for generating better images. In contrast to discriminator-free paradigm training~\cite{li2021revisiting}, we find our DCD empowers the compressed student generator with a better capability to compete against teacher discriminator.
Therefore, we also employ the teacher discriminator to collaboratively determine whether inputs from the student generator are real or not.

%
This work intends to raise the level of feature map distillation to strengthen the compressed student generator to generate high-quality images. The major contributions we have made across the entire paper are listed as follows:
(1) An incentive GAN-oriented discriminator-cooperated feature map distillation method to produce images with high fidelity;
(2) One novel collaborative adversarial training paradigm to better reach a global equilibrium point in compressing GANs;
(3) Remarkable reduction on the generator complexity and significant performance increase.

%

\section{Related Work}
\label{sec:related}

\subsection{GANs and GAN Compression}
Generative adversarial networks (GANs)~\cite{goodfellow2014generative} have attracted the attention of substantive researchers due to their outstanding performance in image generation tasks~\cite{karras2019style, karras2020analyzing, brock2018large, radford2015unsupervised, zhang2019self, isola2017image, zhu2017unpaired, chen2018cartoongan, choi2018stargan}. Since the infancy of GAN, many variants have emerged, from DCGAN~\cite{radford2015unsupervised}, which embraces convolutional neural networks for the first time, to CycleGAN~\cite{zhu2017unpaired} and Pix2Pix~\cite{isola2017image} which implement image-to-image translation, to StyleGAN~\cite{karras2019style} to enable controllable manipulation of various attributes of image synthesis.
CycleGAN~\cite{zhu2017unpaired} intends to transfer a source-domain image into a target style in an unpaired configuration, such as a horse image to a zebra pattern or a summer image to a winter style. On the other hand, Pix2Pix~\cite{isola2017image} is given a corresponding ground-truth image and converts a semantic segmentation or contour map into a photo-realistic picture.
Albeit the performance, GANs suffer heavy burden on storage and computation, which is unfavourable on edge devices.

\begin{figure*}[!t]
  \centering
  \includegraphics[width=0.9\textwidth]{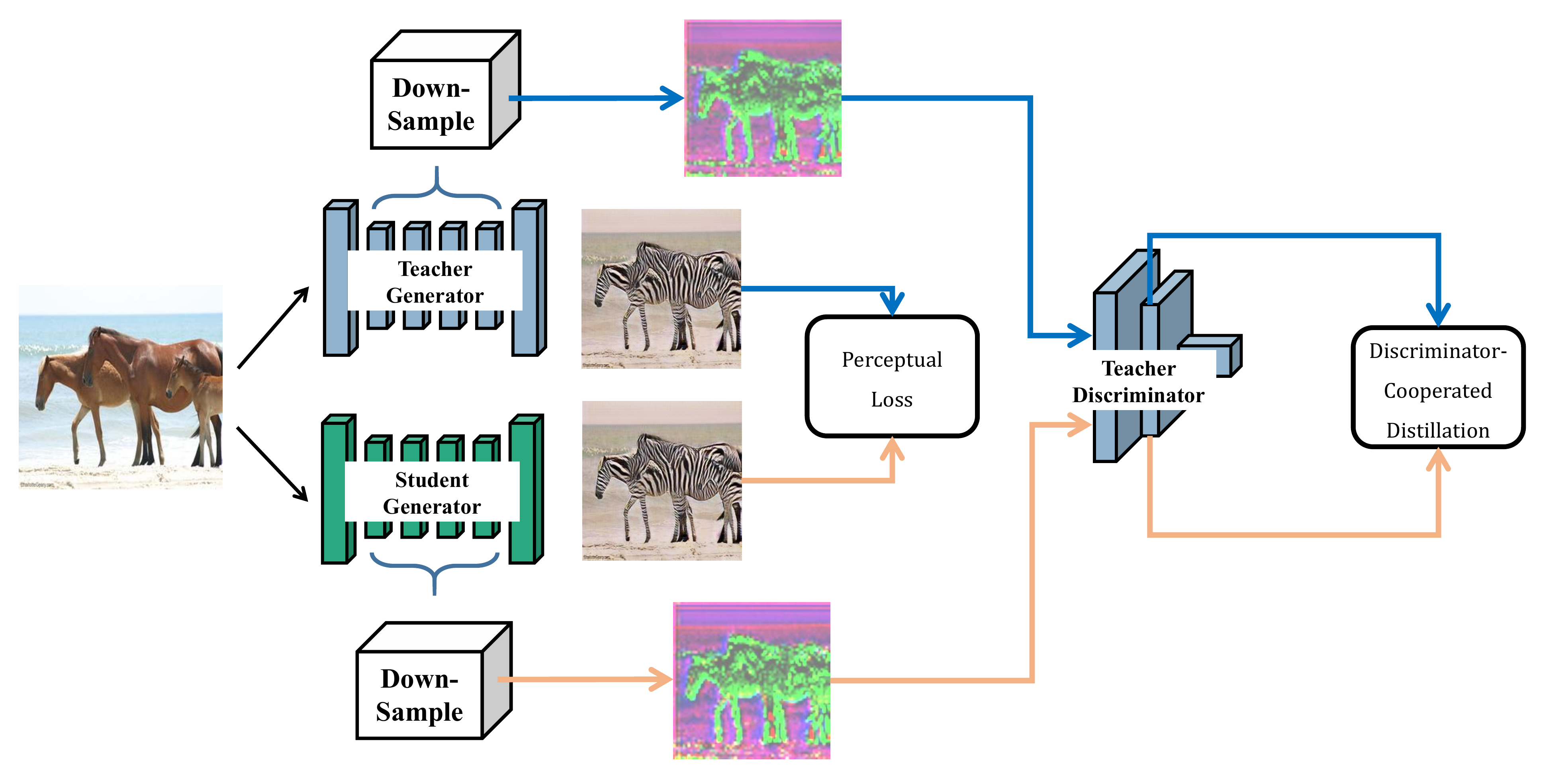}
\vspace{-0.8em}
\caption{Framework of our DCD. Intermediate feature maps from student and teacher are downsampled first to align the dimension, results of which are then fed to the teacher discriminator to minimize distance for a perceptually vivid generated image.}
\vspace{-0.8em}
  \label{fig2: framework}
\end{figure*}

Therefore, recent years have witnessed increasing attention on compressing GANs~\cite{li2022learning, shu2019co, chen2021gans, wang2020gan, wang2019qgan, li2020gan, fu2020autogan, jin2021teachers, kim2022cut, lin2021anycost, zhang2022wavelet, jung2022exploring,you2022exploring}. Co-evolution~\cite{shu2019co} prunes filters in the generator under the constraint of consistent output distributions. 
GAN Slimming~\cite{wang2020gan} suggests a compression framework that can integrate pruning, knowledge distillation and quantization. 
GAN Compression~\cite{li2020gan} views the teacher output as the pseudo label for student generator and unifies the compression framework for GANs trained on paired data and unpaired data. 
DMAD~\cite{li2021revisiting} combines information from discriminator in knowledge distillation, an idea closer to our approach, but the use of discriminator is still under-explored. 
Although the above approaches have compressed parameters and computation costs, both the generated image quality and the model size are still far from practical applications on mobile devices. 
OMGD~\cite{ren2021online} bridges this huge gap in a way by constructing discriminator-free distillation.

\subsection{Knowledge Distillation}
Pioneered by FitNet as of 2014~\cite{romero2014fitnets}, knowledge distillation (KD)~\cite{hinton2015distilling} has become a regular approach in model compression especially to GAN compression, where a larger model with better performance imparts knowledge to a smaller model.
Since then, great efforts have been made to dig out opulent knowledge hints, such as output logits~\cite{jung2022exploring, zhang2022wavelet, zhao2022decoupled}, intermediate feature maps~\cite{romero2014fitnets, chen2021distilling, heo2019comprehensive}, instance relation~\cite{park2019relational,tung2019similarity} and so on.
In this paper, we are mainly inspired by the intermediate feature map based distillation that has been extensively mined to efficiently guide the training of the student network.
Compared to other knowledge hints, feature maps often accommodate a richer level of information and provide more detailed guidance for the student network.
AT~\cite{zagoruyko2016paying} extracts the attention map from the feature maps and trains the student's attention maps to be as close as possible to the one of teacher.
MGD~\cite{yang2022masked} transforms direct learning from the teacher into a generative intermediate goal.
KRD~\cite{chen2021distilling} uses the idea of cross-layer distillation to allow old knowledge from the teacher to guide new knowledge learning from the student network. 
Nevertheless, conventional methods require per-pixel match between feature maps of teacher and student, which does not fit well in GAN compression because the goal of GANs is to generate perceptually similar images.
In this paper, we dig deeper into the distinctiveness of generative adversarial networks.

\section{Methodology}
\label{sec:methodology}

\subsection{Preliminaries}
\label{sec:preliminaries}

Generative adversarial networks~\cite{goodfellow2014generative}, or GANs for short, are an interesting manner to train a generative model by modelling the problem with two sub-models including a generator model $\mathcal{G}$ and a discriminator model $\mathcal{D}$. The two models are trained together in a zero-sum game as:
\begin{equation}
    \begin{split}
    \label{generative adversarial train objective}
        \min\limits_{\mathcal{G}} \max\limits_{\mathcal{D}} \mathcal{L}_{gan} = \mathbb{E}_{y \sim p_{real}} \log \mathcal{D}(y) \\ + \mathbb{E}_{x \sim p(x)}\bigg[\log \Big(1-\mathcal{D}\big(\mathcal{G}(x)\big)\Big)\bigg].
    \end{split}
\end{equation}

Herein, the generator model $\mathcal{G}$ is trained to generate new examples, and the discriminator model $\mathcal{D}$ tries to classify examples as either real (from the domain) or fake (generated). The two models are trained adversarially until the discriminator model is fooled at most times, which indicates that the generator model is producing plausible examples. 
Then, the generator $\mathcal{G}$ is deployed online to complete service for reality.

The serviceability of generator $\mathcal{G}$ rests with not only the performance, but also the hardware capability that makes a greater demand on generator complexity. Therefore, a lighter student generator, $\mathcal{G}^{S}$, can be developed by various methods. The original generator and discriminator, respectively denoted as $\mathcal{G}^T$ and $\mathcal{D}^T$ in this situation to differentiate, play as a teacher to enhance the ability of student generator $\mathcal{G}^S$.

Giving an input variable $x \sim p(x) \in \mathbb{R}^{H \times W \times C}$, we denote the $i$-th layer output of generator as $\mathcal{G}_i(x)$, and $\mathcal{I}_{\mathcal{G}}(x)$ as layer index set of extracted intermediate outputs. Thus, the feature maps based distillation is formulated as:
\begin{equation}
    \label{feature map based distillation}
    \mathcal{L}_{fea-dis} = \sum_{i \in I_\mathcal{G}} \ell\Big( \mathcal{G}_i^T(x) ,  f\big(\mathcal{G}_i^S(x)\big) \Big),
\end{equation}
where ${f(\cdot)}$ is the affine transformation function to align the channel dimensions between the teacher and student, such as 1$\times$1 convolution operators in existing studies~\cite{ren2021online, li2021revisiting, li2020gan}. Also, $\ell(\cdot,\cdot)$ refers to the distance measure function, such as Euclidean distance.

Additionally, perceptual loss~\cite{johnson2016perceptual}, $\mathcal{L}_{per}$, is also widely-adopted in existing studies to encourage natural and perceptually pleasing restored images. $\mathcal{L}_{per}$ comprises a feature reconstruction loss $\mathcal{L}_{fea}$ and a style reconstruction loss $\mathcal{L}_{sty}$ as:
\begin{equation}\label{perceptual_loss}
    \mathcal{L}_{per} = \lambda_{fea} \cdot \mathcal{L}_{fea} + \lambda_{sty} \cdot \mathcal{L}_{sty}.
\end{equation}

Herein, $\mathcal{L}_{fea}$ propels the output representation of the teacher generator to approach to that of the student generator. This is achieved by a pre-trained VGG network $\Phi(\cdot)$~\cite{simonyan2014very} and formalized as:
\begin{equation}
    \label{feature loss}
    \mathcal{L}_{fea} = \sum_{j \in I_{\Phi}}\frac{1}{H_jW_jC_j}\Big\| \Phi_j \big( \mathcal{G}^T(x) \big) - \Phi_j \big( \mathcal{G}^S(x) \big) \Big\|_1,
\end{equation}
where $\Phi_j(\cdot)$ returns the $j$-th activation output of VGG network and $H_j \times W_j \times C_j$ is its shape. Alike to $I_{\mathcal{G}}$, $I_{\Phi}$ contains layer index of extracted intermediate outputs.

As for $\mathcal{L}_{sty}$, it minimizes the difference between Gram matrices of the output and target images in order to preserve style characteristics such as color, textures and common pattern~\cite{ren2021online}. The $\mathcal{L}_{sty}$ is calculated as:
\begin{equation}
    \label{style loss}
    \mathcal{L}_{sty} = \sum_{j \in I_{\Phi}} \Big\| G\Big(\Phi_j \big( \mathcal{G}^T(x) \big)\Big) - G\Big(\Phi_j \big( \mathcal{G}^S(x) \big)\Big) \Big\|_1,
\end{equation}
where $G(\cdot)$ represents abbreviation of Gram matrices.

\subsection{Discriminator-Cooperated Distillation}
\label{sec:method1}
Stepping back and reflecting on the feature map based distillation in Eq.\,(\ref{feature map based distillation}), we realize that a simple utilization of generator capacity is not intact in earlier methods.
The central principle of a GAN is contingent on an ``indirect'' training route through the discriminator updated dynamically to discern how ``realistic'' its input (\emph{i.e.}, generator output) seems. This means that the generator is not trained to minimize the distance from a generated image to a target image, but rather to deceive the discriminator. It is the coopetition pattern between the generator and discriminator that even brings about superficially authentic generated images. Thus, the discriminator is also empowered with informative capacity and must be utilized to enrich the distillation of feature maps.

As shown in Fig.\,\ref{fig2: framework}, we rethink Eq.\,(\ref{feature map based distillation}) and integrate teacher discriminator to cooperate with the distillation process. Alike to feature reconstruction loss defined in Eq.\,(\ref{feature loss}), while taking the generator outputs as its inputs, we accomplish our distillation by aligning the intermediate outputs of the discriminator. We formulate this learning process as:
\begin{equation}
    \label{discriminator-cooperated distillation}
    \mathcal{L}_{dcd} = \sum_{k \in I_\mathcal{D}}\sum_{i \in I_\mathcal{G}} \ell\bigg( \mathcal{D}_k^T\Big( f\big( \mathcal{G}_i^T(x) \big) \Big), \mathcal{D}_k^T\Big( f\big( \mathcal{G}_i^S(x)\big) \Big) \bigg),
\end{equation}
where $\mathcal{D}^{T}_k$ stands for the output of its $k$-th layer. Similar to $I_{\mathcal{G}}$, $I_{\mathcal{D}}$ is a layer index set of the teacher discriminator; herein, $f(\cdot)$ denotes 1$\times$1 convolution operations to downsample the channel dimensions of both the teacher and student feature maps to those of discriminator input. Usually, the channel number of discriminator input is set to three for an RGB-encoded image.
Here, $f(\cdot)$ for teacher generator becomes constant once initialized while that for student generator continues updating for a better fit with teacher.

As an analogy to the vanilla feature map based distillation in Eq.\,(\ref{feature map based distillation}) that compels the per-pixel match between intermediate outputs of both the teacher generator and student generator, the role of the teacher discriminator resembles the pre-trained VGG in Eq.\,(\ref{feature loss}), which acts as a transformed network to enable the intermediate results of student generator to be perceptually alike to these of teacher generator. 
In contrast, it does not require them to do pixel-by-pixel match, since two images might look similar in perspective, but they often have different pixel values, thus we cannot depend on per-pixel match. 
Notice that our discriminator-cooperated distillation is complementary to perceptual loss. The former concentrates on object localization while the latter pays attention to style discrepancies. The analysis of this phenomenon is in Sec.\,\ref{sec:analysis}.

%



\subsection{Collaborative Adversarial Training}
\label{sec:method2}

GANs perform an alternating training paradigm for a unique global equilibrium point: 1) Training the discriminator $\mathcal{D}$ to identify real and generated data while keeping the generator constant; 2) Training the generator $\mathcal{G}$ to generate vivid data that fools the discriminator while keeping the discriminator constant; 3) Repeating steps 1) and 2) till the discriminator model is fooled at most time.
However, the equilibrium is no longer guaranteed when $\mathcal{D}$ and $\mathcal{G}$ are empowered with inconsistent abilities, in which case unstable convergence frequently occurs. 
Generally, the instability stems from two ill-famed issues including vanishing gradient where loss for the generator is zero when the discriminator is perfect, and mode collapse where the stronger generator produces a small set of outputs for any input and the weaker discriminator traps in a local minimum~\cite{arjovsky2017towards}.

Especially, the mode collapse issue is even widespread in GAN compression because the compressed student generator $\mathcal{G}^S$ is powerless to compete with the original full discriminator~\cite{li2022learning}, in particular a pre-trained one~\cite{chen2020distilling, lin2021anycost, jin2021teachers}.
The community has excavated various approaches to weaken the student discriminator. For example, GCC~\cite{li2021revisiting} selectively activates discriminator neurons. Nevertheless, a rule-of-thumb selection has to be carefully designed. Also, training a student discriminator is computationally redundant since it is unwanted in the testing stage.
OMGD~\cite{ren2021online} co-trains teacher generator with the teacher discriminator while the student generator is discriminator-free. The missing adversarial training of student generator somehow barricades the further performance increase.


In this paper, we also reject real student discriminator, and to ensure the teacher's optimization, the teacher discriminator, $\mathcal{D}^T$, is online trained from scratch to be well-matched with the teacher generator $\mathcal{G}^T$. Comparing to ~\cite{ren2021online}, teacher discriminator $\mathcal{D}^T$ also appears in the form of a collaborative discriminator to determine if the inputs from student generator $\mathcal{G}^T$ are real or fake. The major concern originates from the actuality that the teacher discriminator is much powerful than the compressed student generator, which causes mode collapse in adversarial training. Luckily, our discriminator-cooperated distillation in Eq.\,(\ref{discriminator-cooperated distillation}) furnishes student generator with increasing capability to battle against teacher discriminator and leads to better performance than~\cite{ren2021online} as demonstrated in the experiment.


Based on Eq.\,(\ref{generative adversarial train objective}), our adversarial training is rewritten as:
%
%
%
\begin{equation}
    \begin{split}
    \label{trainingT}
        \min\limits_{\mathcal{G}^T,  \mathcal{G}^S} \max\limits_{\mathcal{D}^T} \;&\mathcal{L}_{col} = 
         \mathbb{E}_{y \sim p_{real}} \log \mathcal{D}^T(y) 
        \\& + \mathbb{E}_{x \sim p(x)}\bigg[\log \Big(1-\mathcal{D}^T\big(\mathcal{G}^T(x)\big)\Big)\bigg] 
        \\& + \lambda_{stu} \cdot \mathbb{E}_{x \sim p(x)}\bigg[\log \Big(1-\mathcal{D}^T\big(\mathcal{G}^S(x)\big)\Big)\bigg],
    \end{split}
\end{equation}
where $\lambda_{stu}$ refers to a trade-off parameter; and $\lambda_{stu} = 0$ degenerates to discarding student discriminator when co-trained with the student generator~\cite{ren2021online}.

\subsection{Training Objective}
\label{sec:loss}
Looking back to Sec.\,\ref{sec:preliminaries}, the loss terms in most conventional feature map based distillation methods include $\mathcal{L}_{gan}$ in Eq.\,(\ref{generative adversarial train objective}), $\mathcal{L}_{fea-dis}$ in Eq.\,(\ref{feature map based distillation}) and $\mathcal{L}_{per}$ in Eq.\,(\ref{perceptual_loss}). In this paper, we improve $\mathcal{L}_{fea-dis}$ through a discriminator-cooperated feature map distillation loss $\mathcal{L}_{dcd}$ in Eq.\,(\ref{discriminator-cooperated distillation}), and $\mathcal{L}_{gan}$ through collaborative adversarial training loss $\mathcal{L}_{col}$ in Eq.\,(\ref{trainingT}). Therefore, the overall training objective in this paper is given in the following:
\begin{equation}\label{objective}
    \min\limits_{\mathcal{G}^T,  \mathcal{G}^S} \max\limits_{\mathcal{D}^T}\, (\mathcal{L}_{gan}  + \mathcal{L}_{per} + \lambda_{dcd} \cdot \mathcal{L}_{dcd}),
\end{equation}
where $\lambda_{dcd}$ balances the loss term. Four hyper-parameters: $\lambda_{fea}$, $\lambda_{sty}$, $\lambda_{stu}$ and $\lambda_{dcd}$ are used in this paper, influence of each parameter is ablated in the appendix.

\section{Experimentation}
\label{sec:experiment}

%

\begin{table*}[!t]
    \caption{Performance comparison when compressing CycleGAN on horse2zebra, and Pix2Pix on edges2shoes. }
    \vspace{-1.5em}
    \label{tab:table1}
  \begin{center}
    \begin{tabular}{c|c|c|c|c|c}    
    \toprule
      \textbf{Model} &\textbf{Dataset} & \textbf{Method} & \textbf{MACs} & \textbf{\#Parameters} & \textbf{FID($\downarrow$)} \\
      \hline
      \multirow{15}*{CycleGAN~\cite{zhu2017unpaired}} &
      \multirow{8}*{horse2zebra~\cite{zhu2017unpaired}}
      & Original~\cite{zhu2017unpaired} & 56.80G(1.0$\times$) & 11.30M(1.0$\times$) & 61.53 ({\color{blue}-})\\
      & & Co-Evolution~\cite{shu2019co} & 13.40G(4.2$\times$) & - & 96.15({\color{blue}-34.62}) \\
      & & DMAD~\cite{li2022learning} & 2.41G(23.6$\times$) & 0.28M(40.0$\times$) & 62.96({\color{blue}-1.43}) \\
      & & Wavelet KD~\cite{zhang2022wavelet} & 1.68G(33.8$\times$) & 0.72M(15.81$\times$) & 77.04({\color{blue}-15.51})  \\
      & & GAN-Compression~\cite{li2020gan} & 2.67G(21.3$\times$) & 0.34M(33.2$\times$) & 64.95({\color{blue}-3.42})  \\
      & & GCC~\cite{li2021revisiting} & 2.40G(23.6$\times$) & - & 59.31({\color{red}+2.22})  \\
      & & OMGD~\cite{ren2021online} & 1.408G(40.3$\times$) & 0.137M(82.5$\times$) & 51.92({\color{red}+9.61}) \\
      & & \textbf{DCD} (Ours) & \textbf{1.408G(40.3$\times$)} & \textbf{0.137M(82.5$\times$)} & \textbf{48.24({\color{red}+13.29})}  \\
      \cline{2-6}
      &\multirow{6}{*}{summer2winter~\cite{zhu2017unpaired}} 
      & Original~\cite{zhu2017unpaired} & 56.80G(1.0$\times$) & 11.30M(1.0$\times$) & 79.12({\color{blue}-})\\
      & & Co-Evolution~\cite{shu2019co} & 11.10G(5.1$\times$) & - & 78.58({\color{red}+0.54})\\
      & & AutoGAN-Distiller~\cite{fu2020autogan} & 4.34G(13.1$\times$) & - & 78.33({\color{red}+0.79})  \\
      & & DMAD~\cite{li2022learning} & 3.18G(17.9$\times$) & 0.30M(37.7$\times$) & 78.24({\color{red}+0.88})  \\
      & & OMGD~\cite{ren2021online} & 1.408G(40.3$\times$) & 0.137M(82.5$\times$) & 73.79({\color{red}+5.33}) \\
      & & \textbf{DCD} (Ours) & \textbf{1.408G(40.3$\times$)} & \textbf{0.137M(82.5$\times$)} & \textbf{73.63({\color{red}+5.49})}  \\
      \hline
      \multirow{6}*{Pix2Pix~\cite{isola2017image}} &
      \multirow{5}*{edges2shoes}
      & Original~\cite{isola2017image} & 18.60G(1.0$\times$) & 54.40M(1.0$\times$) & 34.31({\color{blue}-}) \\
      & & DMAD~\cite{li2022learning} & 2.99G(6.2$\times$) & 2.13M(25.5$\times$) & 46.95({\color{blue}-12.64})  \\
      & & Wavelet KD~\cite{zhang2022wavelet} & 1.56G(11.92$\times$) & 13.61M(4.00$\times$) & 80.13({\color{blue}-45.82}) \\
      & & OMGD~\cite{ren2021online} & 1.219G(15.3$\times$) & 3.404M(16.0$\times$) & 25.00({\color{red}+9.41}) \\
      & & \textbf{DCD} (Ours) & \textbf{1.219G(15.3$\times$)} & \textbf{3.404M(16.0$\times$)} & \textbf{23.43({\color{red}+10.98})}  \\
      \bottomrule
    \end{tabular}
  \end{center}
\vspace{-1.5em}
\end{table*}

\subsection{Setups}
\textbf{GAN Models and Benchmarks}. 
We present the performance of compressed CycleGAN~\cite{zhu2017unpaired} and Pix2Pix~\cite{isola2017image} to follow and compare with existing methods~\cite{li2022learning, shu2019co, chen2021gans, wang2020gan,  li2020gan, jin2021teachers, zhang2022wavelet, jung2022exploring}. For a fair comparison with the current SoTA OMGD~\cite{ren2021online}, the compressed generators (students) consist of 1/4 channels of the original full ResNet generators (teachers)~\cite{li2020gan}.
CycleGAN translates images from one domain to another without a one-to-one mapping between the source and target domain. Therefore, we verify the performance upon unpaired image translation benchmarks including horse2zebra~\cite{zhu2017unpaired} and summer2winter~\cite{zhu2017unpaired}.
As for Pix2Pix, we perform distillation upon paired edges2shoes~\cite{Yu_2014_CVPR} because it requires learning a mapping from input images to output images.

\textbf{Evaluations}.
Fr$\Acute{\mathbf{e}}$chet Inception Distance (FID), or FID for short~\cite{NIPS2017_8a1d6947}, is particularly developed to evaluate the performance of GANs. It accesses the quality of generated images by an Inception-V3 network~\cite{Szegedy_2016_CVPR} to separately embed synthetic and real images to feature space, and then calculate the Wasserstein distance of their distributions. The lower FID scores indicate the better quality of generated images.


%
\textbf{Implementations}.
We train CycleGAN and Pix2Pix for a total of 100 epochs. The initial learning rate is given as 2$\times$$e$$^{\text{-4}}$ and then linearly decayed to 0 as training goes. The batch size is set to 4 on edges2shoes, and 1 on horse2zebra~\cite{zhu2017unpaired} and summer2winter~\cite{zhu2017unpaired}. 
We have $\lambda_{dcd}$=1, $\lambda_{fea}$=1$\times$$e$$^{\text{1}}$, $\lambda_{sty}$=1$\times$$e$$^{\text{4}}$ and $\lambda_{stu}$=1 across all experiments. In the appendix, specific ablations in regard to these hyper-parameters have been provided.

\subsection{Quantitative Comparison}
\textbf{GAN Compression}.
We first compare with existing implementations on GAN compression in Table\,\ref{tab:table1} where the {\color{blue}blue} number indicates performance drop, while the {\color{red}red} number denotes performance increase compared with the original GAN models.
We conclude the following observations from Table\,\ref{tab:table1}: 
First, on summer2winter, all methods lead to a performance increase of more or less. On the contrary, most methods cause FID drops on the challenging horse2zebra and edges2shoes while methods such as OMGD~\cite{ren2021online} and our DCD consistently enhance the performance.
Second, our DCD leads to the most complexity reduction \emph{w.r.t.} MACs and parameters, meanwhile it gains the best performance increase on all the three benchmark datasets. 
Third, with the same reduction of MACs (40.3$\times$) and parameters (82.5$\times$) on unpaired CycleGAN, our DCD well outperforms the recent SoTA method, \emph{i.e.}, OMGD. In particular, we drastically increase the performance of 51.92 for OMGD to 48.24, which is also 13.29 increase compared to the original CycleGAN. This increase is very challenging since 51.92 already is a strong performance. Nevertheless, our increase is very evident.
Lastly, similar to CycleGAN, the results on paired Pix2Pix also show the superiority of our DCD over OMGD when compressing 15.3$\times$ MACs and 16.0$\times$ parameters. In this case, OMGD leads to 9.41 FID gains while our DCD has a better increase of 10.98.

Therefore, our discriminator-cooperated feature map distillation has well demonstrated its great capability to boost the performance of a light-weighted generator.

\begin{table}[!t]
 \centering
 \caption{Comparison between different feature map distillation methods on horse2zebra.}
 \vspace{-0.8em}
 \begin{tabular}{c|c|c}\toprule
    \textbf{Dataset} &\textbf{Method} & \textbf{FID($\downarrow$)} \\
    \hline
    \multirow{5}{*}{horse2zebra~\cite{zhu2017unpaired}} 
    & Baseline & 65.13 \\
    & FitNet~\cite{romero2014fitnets} & 65.84\\
    & MGD~\cite{yang2022masked} & 67.57 \\
    & KRD~\cite{chen2021distilling} & 61.53\\
    \cline{2-3}
    & \textbf{DCD} (Ours) & \textbf{48.24} \\
    \bottomrule
 \end{tabular}
 \label{exp: ablation 1}
 \vspace{-1.5em}
\end{table}

\textbf{Feature Map Distillation}.
%
As introduced in Sec.\,\ref{sec:intro}, current feature map distillation methods such as FitNet~\cite{romero2014fitnets}, MGD~\cite{yang2022masked} and KRD~\cite{chen2021distilling} construct pixel-to-pixel matching between the student generator and teacher generator. In contrast, our DCD is constrained to generate perceptually alike images. In Table\,\ref{exp: ablation 1}, we replace our DCD with the aforementioned distillation scenarios for performance comparison. All experiments are performed on CycleGAN of 40.3$\times$ MACs reduction as shown in Table\,\ref{tab:table1}. We can see that FitNet and MGD incur performance degradation of 0.71 and 2.44 FID. Though KRD allows cross-layer connections and rises the FID by 4.40, the performance increase is very limited if compared with our 13.29 performance gains. To find out the root cause, these methods were initially invented to perform image classification that focuses more on extracting robust image feature vectors. However, GANs often pay great attention to the image contents, which do not call for identical pixel values between two images but urge more for perceptual discrepancies. Therefore, off-the-shelf per-pixel matching studies fail to fit well when directly extended to GAN compression.

\begin{figure*}[!t]
  \centering
  \includegraphics[width=0.9\textwidth]{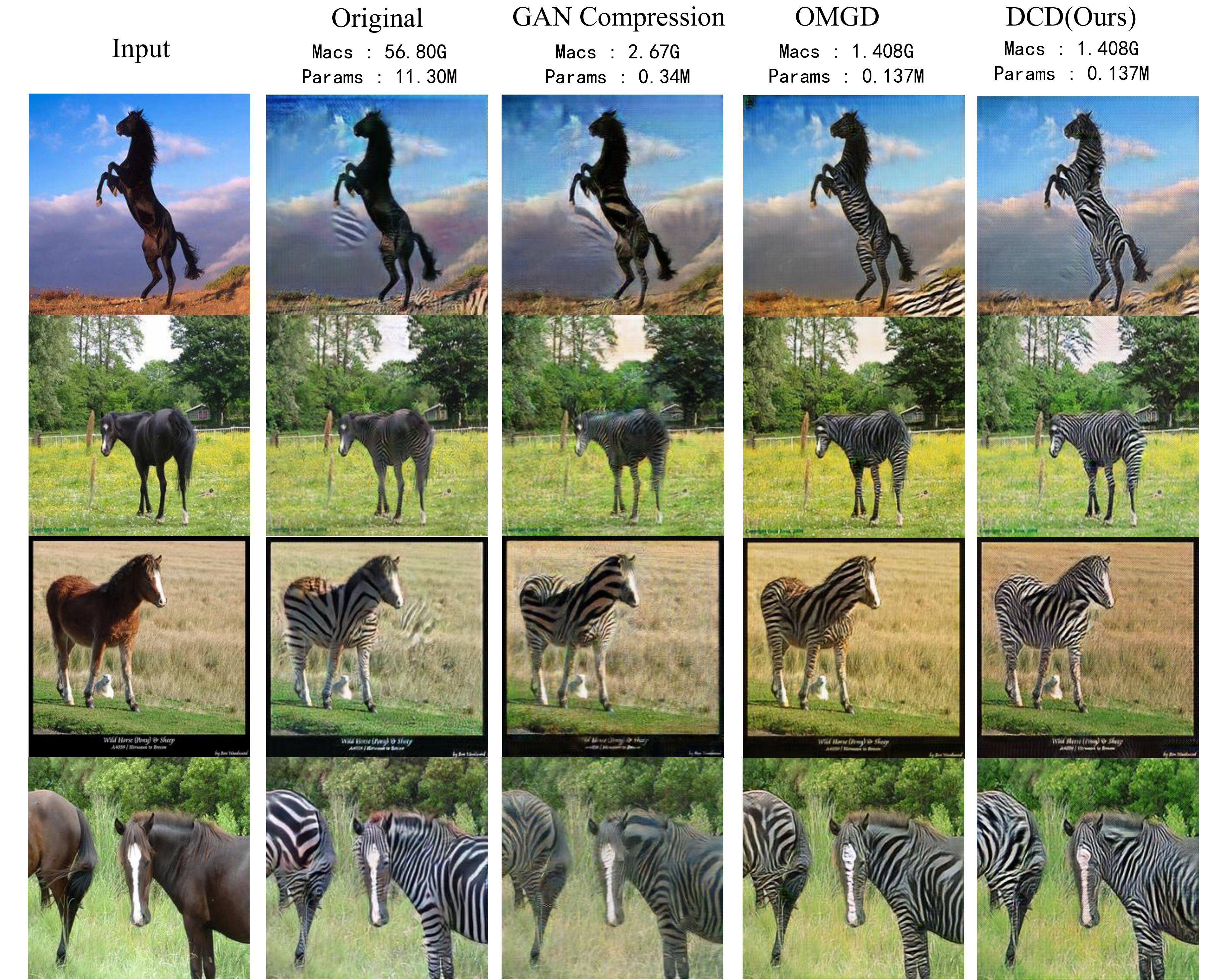}
\vspace{-0.5em}
\caption{Visualization comparison on horse2zebra with CycleGAN.}
  \label{fig3: visual results}
\vspace{-1.5em}
\end{figure*}

\subsection{Visualization}
We further present visualization of generated images from original CycleGAN generator as well as its compressed versions in Fig\,\ref{fig3: visual results}. Evidently, our DCD results in not only more vivid objects (zebras), but also well retains background information. Regarding the objects, our DCD produces sharper and brighter stripes compared to other methods. In particular, existing methods are more or less influenced by the input. For example, the brown fur on horse is retained on the generated zebras (third row) while DCD well overcomes this drawback. As for the background, we find that DCD sometimes presents better visual perception even than the inputs, such as greener meadows and trees (last row). 
The better visual results are in accordance with better FID in Table\,\ref{tab:table1}. 
More examples of summer2winter and edges2shoes can be referred to the appendix.


\subsection{Analysis}
\label{sec:analysis}
We continue to conduct ablation studies to analyze influences of different training losses defined in Eq.\,(\ref{objective}), as well as our downsampling strategies defined in Eq.\,(\ref{discriminator-cooperated distillation}), trying to reveal why our DCD performs well. All experiments are constructed by using CycleGAN on horse2zebra.

\begin{table}[!t]
\centering
\caption{Training loss influence to CycleGAN on horse2zebra.}
\vspace{-0.5em}
\begin{tabular}{c|c|c|c}\toprule
    $\mathcal{L}_{per}$ &$\mathcal{L}_{dcd}$ &$\mathcal{L}_{gan}$ & \textbf{FID($\downarrow$)}\\
    \hline
    \checkmark &  &  & 67.20\\
    & \checkmark &  & 324.27 \\
     &  &  \checkmark& 389.77\\
    \checkmark & \checkmark &  & 55.24\\
    \checkmark &  & \checkmark & 65.13\\
     & \checkmark & \checkmark & 323.63\\
    \checkmark & \checkmark & \checkmark & \textbf{48.24}\\
    \bottomrule
\end{tabular}
\label{exp: ablation 3}
\vspace{-1.5em}
\end{table}

\textbf{Training Loss}. 
Table\,\ref{exp: ablation 3} manifests the performance of different loss combinations. We observe the significance of the perceptual loss, without which the FID drastically increases to hundreds. Also, both our DCD loss and adversarial training loss are complementary to perceptual loss where $\mathcal{L}_{per}$+$\mathcal{L}_{dcd}$ increases performance to 55.24 and it is 65.13 for $\mathcal{L}_{per}$+$\mathcal{L}_{gan}$.
Combining all the training losses results in the optimal FID of 48.24.


\begin{figure*}[!t]
  \centering
  \includegraphics[width=0.99\textwidth]{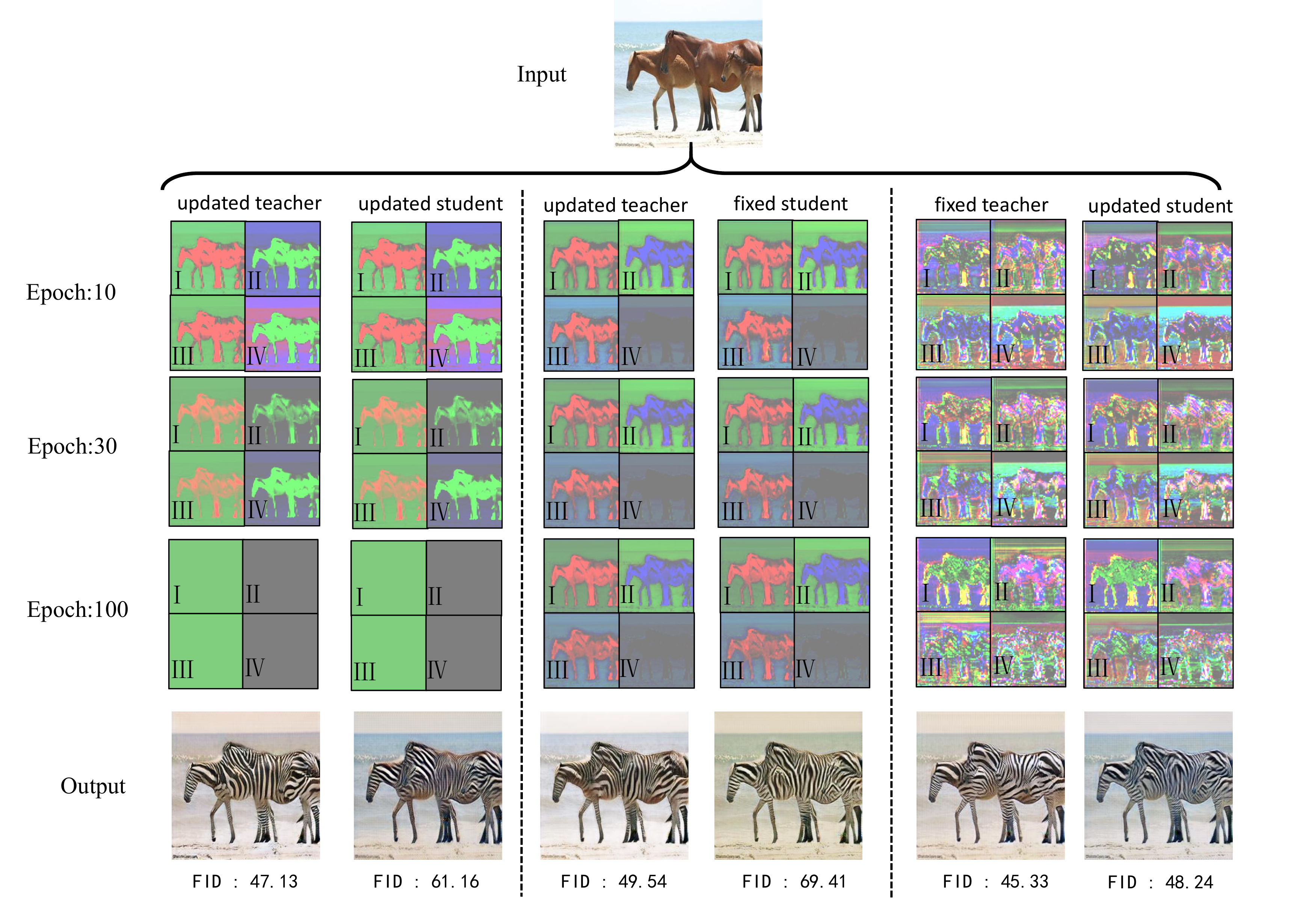}
  \vspace{-1.5em}
\caption{Feature map visualization of teacher and student generators at different training stages.}
\vspace{-1.0em}
  \label{fig5: feature ablation}
\end{figure*}

\textbf{Downsampling}.
In Eq.\,(\ref{discriminator-cooperated distillation}), we adopt 1$\times$1 convolution operations to downsample feature maps to a three-channel RGB-encoded image before being fed to the teacher discriminator.
We fix downsampling modules for teacher generator while updating those for student generator, which is very crucial to the performance of the student generator. We visualize RGB features from layer 3, 6, 9, and 12 of the teacher and student generators at different training stages, referred to as stage \uppercase\expandafter{\romannumeral1}$\sim$\uppercase\expandafter{\romannumeral4} in order. Results are displayed in Fig.\,\ref{fig5: feature ablation} where two variants are introduced for comparison including updating both the teacher and student downsampling modules, and updating the teacher downsampling modules while fixing those of student.
We can observe that the student generator keeps pace with teacher generator, so that they generate similar feature maps. This indicates the teacher knowledge has been well transferred to learn student model. 
Unfortunately, updating the teacher downsampling modules severely damages the knowledge from the teacher generator where only 61.16 FID is obtained if the student downsampling modules are updated as well and 69.41 otherwise. 
The poor performance mainly stems from the broken feature maps of teacher, where horse objects become invisible. Therefore, the student generator collapses as well.
As for our downsampling strategy where downsampling modules are fixed for teacher generator while updated for student generator, the attention is gradually paid to the target horse objects as network training.
Therefore, our discriminator-cooperated feature map distillation benefits more to localizing target objects. This is complementary to the perceptual loss that is engaged in style transferring.
Their combination leads the lightweight student generator to finally synthesise vivid images with intricate textures.
\section{Conclusion}
\label{sec:conclution}
In this paper, we proposed a novel discriminator-cooperated distillation (DCD) that transfers feature maps knowledge from teacher generator, to optimize the compressed student generator for perceptually better image generation in an economical fashion. Our DCD reshapes the conventional pixel-to-pixel feature map match by skillfully utilizing the teacher discriminator as a transformation to pursue better visual perception in generated images. Our methods show that the teacher discriminator can also be utilized to co-train with the compressed student generator and accordingly invent a collaborative adversarial training paradigm.
Our experimental results demonstrated the significant improvement of DCD in both quantitative and qualitative performance, meanwhile, the complexity of student generator is reduced by a large volume.

\section*{Acknowledgement}
This work is supported by the National Science Fund for Distinguished Young (No.62025603), the National Natural Science Foundation of China (No.62025603, No. U1705262, No. 62072386, No. 62072387, No. 62072389, No, 62002305, No.61772443, No. 61802324 and No. 61702136) and Guangdong Basic and Applied Basic Research Foundation (No.2019B1515120049).

{\small
\bibliographystyle{ieee_fullname}
\bibliography{main}
}

\clearpage

\section*{Appendix \label{appendix}}

\begin{figure*}[!t]
  \centering
  \includegraphics[width=0.95\textwidth]{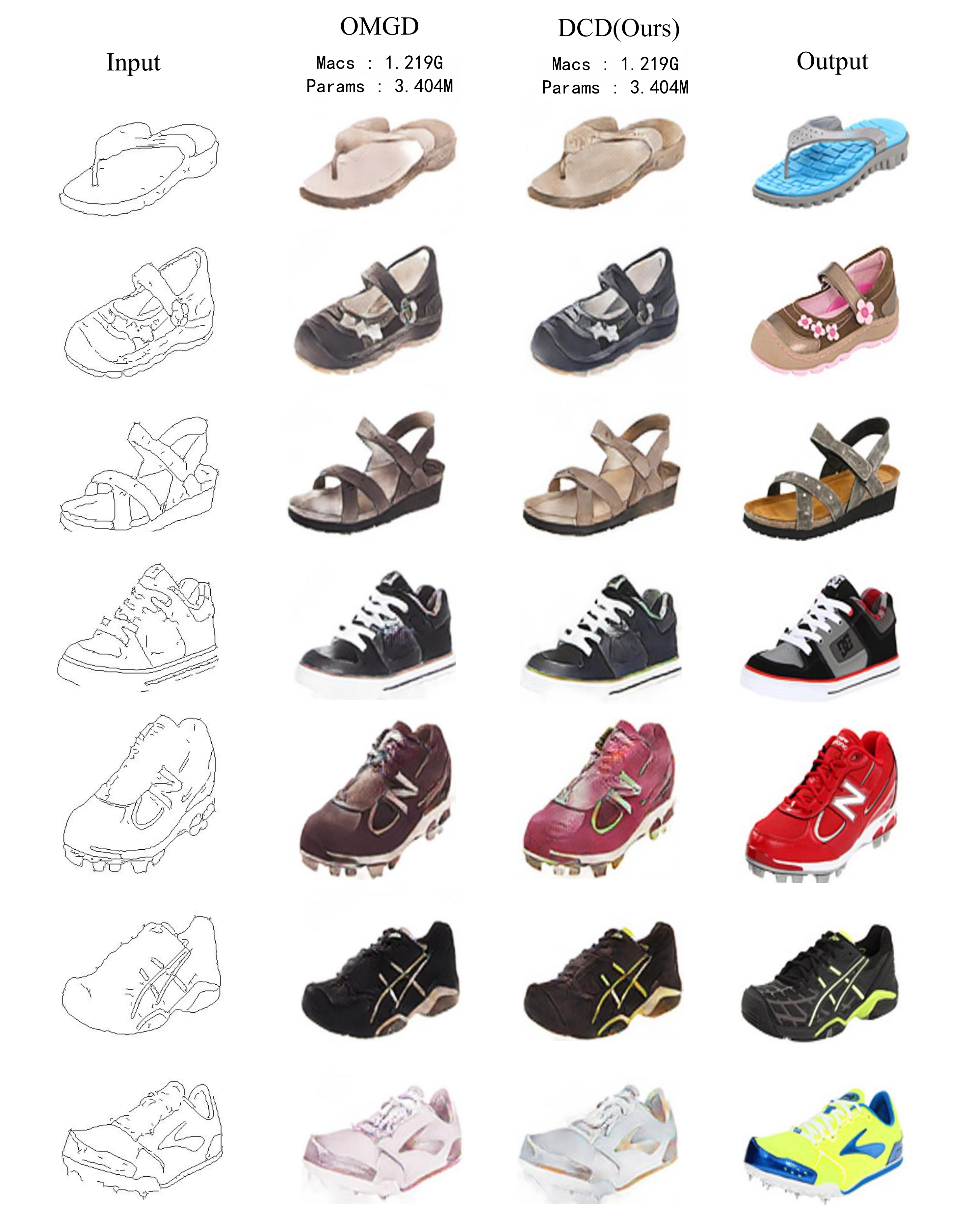}
\caption{Visualization comparison on edges2shoes with Pix2pix.}
  \label{fig1: edges2shoes}
\end{figure*}

\begin{figure*}[!t]
  \centering
  \includegraphics[width=1.0\textwidth]{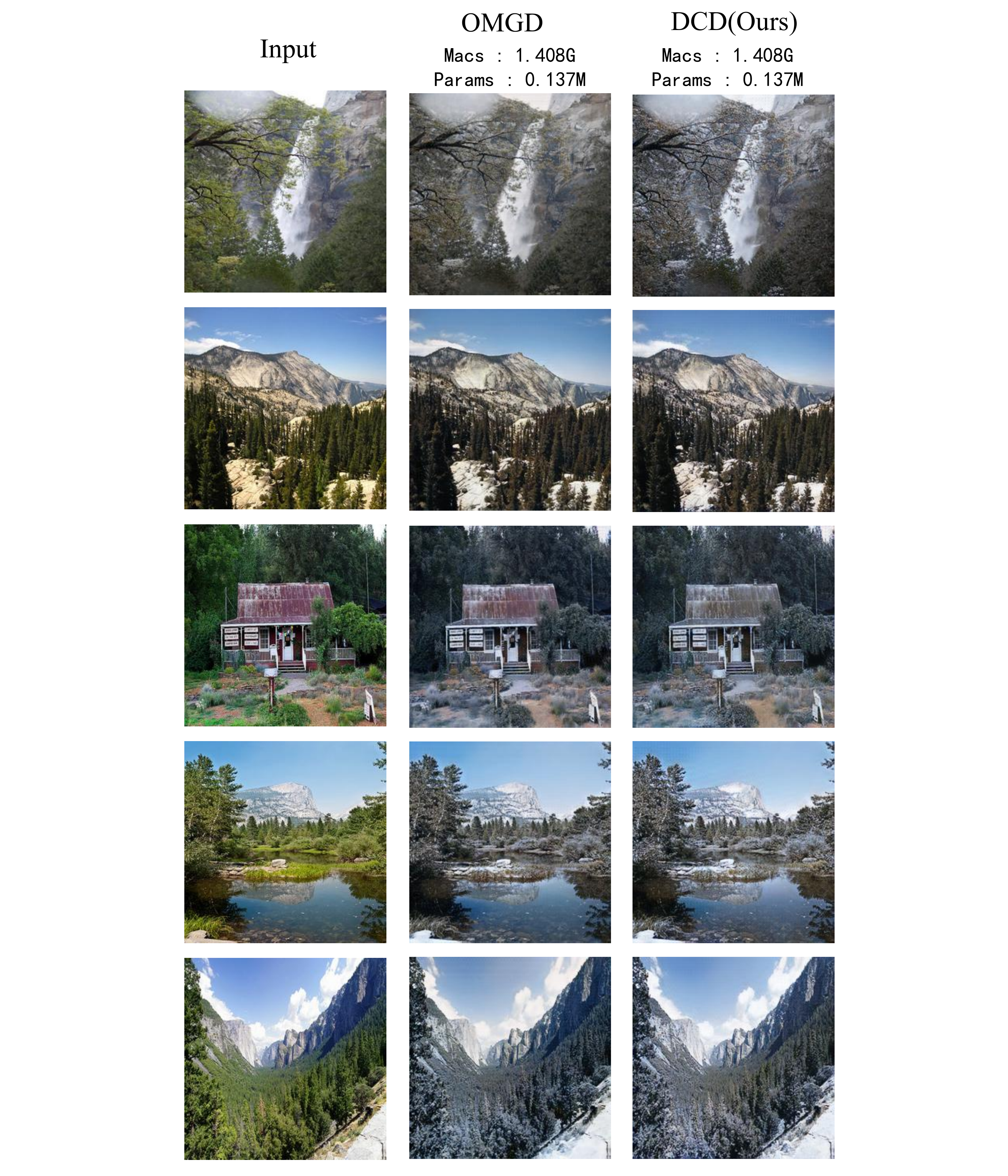}
\caption{Visualization comparison on summer2winter with CycleGAN.}
  \label{fig2: summer2winter}
\end{figure*}

\section{Additional Ablation Study}

\begin{table}[htbp]
 \centering
  \caption{Different values of $\lambda_{dcd}$ in horse2zebra.}
 \label{tab:lambda_dcd}
 \begin{tabular}{c|c}\toprule
     \textbf{$\lambda_{dcd}$} & \textbf{FID($\downarrow$)} \\
    \hline
    0.1 & 58.76 \\
    5 & 58.514 \\
    10 & 59.94 \\
    \cline{1-2}
    \textbf{1 (Ours)} & \textbf{48.24} \\
    \bottomrule
 \end{tabular}
\end{table}

\textbf{Trade-off Parameter of Discriminator-cooperated Distillation (DCD)}. Table~\ref{tab:lambda_dcd} shows the ablation experiments on horse2zebra using CycleGAN with the various trade-off value $\lambda_{dcd}$. In the table the FID increase is limited if the hyperparameter is large or small, whereas the best results can be obtained as $\lambda_{dcd}$ is set to 1.

\begin{table}[htbp]
 \centering
  \caption{Different values of $\lambda_{stu}$ in horse2zebra.}
 \label{tab:lambda_stu}
 \begin{tabular}{c|c}\toprule
     \textbf{$\lambda_{stu}$} & \textbf{FID($\downarrow$)} \\
    \hline
    0.1 & 69.97 \\
    10 & 60.73 \\
    \cline{1-2}
    \textbf{1 (Ours)} & \textbf{48.24} \\
    \bottomrule
 \end{tabular}
\end{table}

\textbf{Trade-off Parameter of Adversarial Training}. Considering the teacher discriminator as an alternative model to the student discriminator, we adjusted that weight $\lambda_{stu}$ to see its impact. As seen in Table~\ref{tab:lambda_stu}. The negative effect is brought about when the weight $\lambda_{stu}$ is too large, i.e., set to 10, which drops to 60.73; while reducing the weight causes an even greater drop in performance, which drops the FID to 69.97. The optimal performance of 48.24 is obtained by setting the trade-off value $\lambda_{stu}$ to 1.

\section{Additional Visualization}
In this section, we supplement the visual comparison images of edges2shoes and summer2winter.

\textbf{Edges2shoes}.
The edges2shoes task refers to filling a shoe's outline sketch with a shoe that contains rich patterns and colors, which can be done using Pix2Pix. We compared the previous state-of-the-art method, namely OMGD, with the DCD used in this paper, demonstrating in various ways that our method can render more colorful and clear for the generation of detailed textures. As the images in the fifth and sixth rows of Fig.\,\ref{fig1: edges2shoes} show, DCD is more sensitive to colors, such as green and red and tends to produce brighter and sharper textures in the generated results, which leads to more vibrant generation effects.

\textbf{Summer2winter}.
The summer2winter task refers to adding snow, blue light and other winter-specific effects to input summer images, making the images rendered with a winter mood. In Fig.\,\ref{fig2: summer2winter}, the overall generation effects of DCD and OMGD on summer2winter are relatively close, yet DCD is more sensitive to vivid colors and can generate more vivid winter landscapes. The various generated results, prove that DCD can slightly more focus on blue tones so that the generated winter images have a more wintery feel.

\end{document}